\documentclass[letterpaper, 11pt]{article}
\usepackage{graphicx}
\usepackage{longtable}
\usepackage{wrapfig}
\usepackage{rotating}
\usepackage[normalem]{ulem}
\usepackage{amsmath}
\usepackage{amssymb}
\usepackage{capt-of}
\usepackage{hyperref}
\usepackage[table, dvipsnames]{xcolor}
\usepackage{listings}
\usepackage{url}
\usepackage{amsthm}
\usepackage{amsfonts}
\usepackage{tcolorbox}
\usepackage{graphicx}
\usepackage{wrapfig}
\usepackage{listings}
\usepackage{framed}
\usepackage{float}
\usepackage{tikz}
\definecolor{navyblue}{HTML}{000080}
\hypersetup{colorlinks=true, citecolor=RoyalBlue, linkcolor=BrickRed, urlcolor=black} % by default sets linkcolor=BrickRed, citecolor=green, filecolor=cyan, menucolor=red, runcolor=cyan, urlcolor=magenta
\usepackage{booktabs}
\usepackage{longtable}
\usepackage{array}
\usepackage{multirow}
\usepackage{hhline}
\usepackage{caption}
\usepackage{diagbox}
\makeatletter\newcommand\arraybslash{\let\\\@arraycr}\makeatother
\usepackage{enumitem}
\setlist[description]{leftmargin=0pt}
\setlist[itemize]{itemsep=0pt}
\setlist[enumerate]{noitemsep}
\theoremstyle{definition}

\theoremstyle{definition}

\theoremstyle{definition}

\theoremstyle{definition}

\theoremstyle{definition}

\usepackage[left=2.75cm,right=2.75cm,top=3cm,bottom=4cm]{geometry}
\usepackage[round]{natbib}\renewcommand{\cite}{\citep} % add \bibliography{../readings/sap}
\author{Partha Ghosh\\University of Tübingen}
\usepackage{setspace}
\usepackage[parfill]{parskip}
\date{}
\title{Non-watertight Mesh Reconstruction}
\hypersetup{
 pdfauthor={},
 pdftitle={Non-watertight Mesh Reconstruction},
 pdfkeywords={},
 pdfsubject={},
 pdfcreator={Emacs 28.1 (Org mode 9.5.2)}, 
 pdflang={English}}
\begin{document}

\maketitle
\begin{abstract}
\noindent Reconstructing 3D non-watertight mesh from an unoriented point cloud is an unexplored area in computer vision and computer graphics. In this project, we tried to tackle this problem by extending the learning-based watertight mesh reconstruction pipeline presented in the paper ‘Shape as Points’. The core of our approach is to cast the problem as a semantic segmentation problem that identifies the region in the 3D volume where the mesh surface lies and extracts out the surfaces from the detected regions. Our approach achieves compelling results compared to the baseline techniques.
\end{abstract}

\section{Introduction}
\label{sec:org4eee532}
Reconstructing surfaces from scanned 3D points has been an important research
area for several decades. With the wide proliferation of 3D scanners, the
problem of surface reconstruction has received significant attention in the
graphics and vision communities.  In recent years, neural implicit
representations gained popularity in 3D reconstruction due to their
expressiveness and flexibility. But, the existing works in this area
concentrated on the representation and reconstruction of watertight mesh only.

In computer vision and graphics, watertight meshes usually describe meshes
consisting of one closed surface. In this sense, watertight meshes do not
contain holes and have a clearly defined inside \cite{enwiki:1024744141}. But
in reality, most of the objects are not watertight and therefore the
reconstruction of the mesh from the point cloud sampled from such objects
should also reflect the non-watertightness. But reconstructing non-watertight
meshes remained an unexplored area in this domain. Therefore, in this project,
we tried to tackle this problem by extending the learning-based 3D watertight
mesh reconstruction pipeline presented in the paper ‘Shape as Points’ (SAP)
\cite{peng2021shape}.

The existing pipeline in SAP can only be used to reconstruct the watertight
mesh from unoriented point cloud even when the point cloud exhibits
non-watertightness. Therefore to achieve our goal, it is required to detect
and extract out only the relevant part of the watertight mesh defined by the
point cloud.

The core of our approach is to cast the problem as a semantic segmentation
problem. We take the output representation from the SAP pipeline and apply
semantic segmentation to it to identify the region in the 3D volume where the
mesh surface lies and then we can extract the non-watertight mesh with the
application of marching cube algorithm. The advantage of our approach is its
simplicity and robustness. Compared to the hand-engineered filtering
techniques that we used as baselines, our method achieves compelling results
both qualitatively and quantitatively.

In summary, the main contributions of this work are:
\begin{itemize}
\item We present a novel machine learning-based approach for generating
high-quality non-watertight meshes.
\item We show that our approach achieves significantly better results than the
hand-engineered filtering based baseline methods both qualitatively and
quantitatively.
\end{itemize}

We organize the structure of the report as follows. We first provide an
overview of the learning-based pipeline in SAP and its limitations in Section
\ref{org2f64bb1}. We then introduce details of our methodology in Section \ref{orga7e8a2d}, followed
by the description of the baseline methods and the effectiveness of our
proposed model compared to the baselines both quantitatively and qualitatively
in Section \ref{orge45565e}.

\section{Reviewing SAP \label{org2f64bb1}}
\label{sec:orgf519865}

In this section, we briefly review the learning-based watertight surface
reconstruction pipeline in SAP \cite{peng2021shape} and analyze the limitation
of PSR indicator grid for non-watertight mesh reconstruction.

\subsection{Learning-based Watertight Surface Reconstruction}
\label{sec:orgbf95f47}

\begin{figure}
\centering
\includegraphics[width=\textwidth]{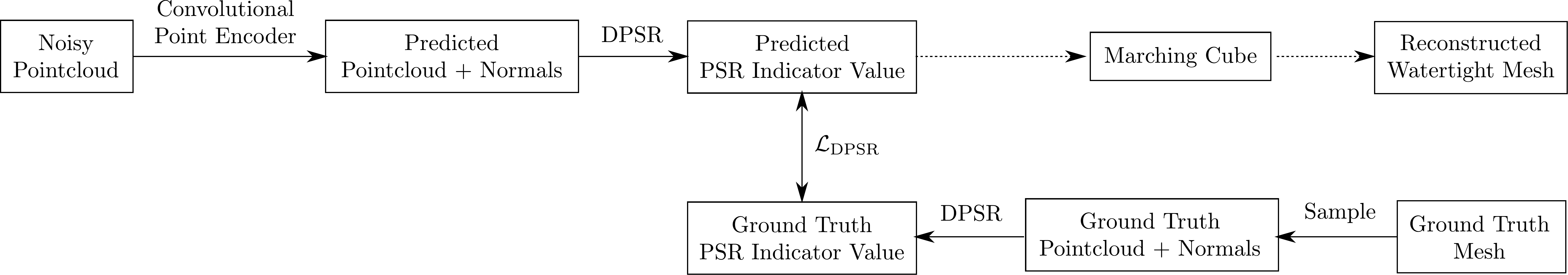}
\caption{Pipeline for learning-based watertight surface reconstruction in SAP}
\label{sap_pipeline}
\end{figure}

The learning-based watertight surface reconstruction setting in SAP takes a
noisy, unoriented point cloud as input and outputs a watertight mesh. More
specifically, taking the noisy, unoriented point cloud as input the network
predicts a clean oriented point cloud which is then fed into the
Differentiable Poisson Solver (Section \ref{org7ec7f79}) to produce an occupancy indicator
grid and the watertight mesh is then extracted by running the Marching Cube
algorithm on this occupancy indicator grid. The key idea of this work was to
introduce differentiability in the classic Poisson Surface Reconstruction
algorithm. The model was trained with watertight meshes as ground truth and
consequently was supervised directly with the ground truth occupancy grid
obtained from these meshes. Figure \ref{sap_pipeline} illustrates the
pipeline of the learning-based surface reconstruction task.

\subsubsection{Differentiable Poisson Solver \label{org7ec7f79}}
\label{sec:orgea8fa9f}
The Differentiable Poisson Solver \cite{peng2021shape} is the
differentiable version of the classic Poisson Surface Reconstruction (PSR)
algorithm \cite{Kazhdan2006}. The PSR algorithm constructs the
characteristic function \(\chi\) of the solid defined by the oriented point
cloud --- the function whose value is one inside of the solid and zero
outside of it --- and then extracts the appropriate iso-surface. The
characteristic function when realized in the voxel grid, produces the
PSR indicator grid.

Let \(\mathcal{P}=\left\lbrace (\textbf c_{i}, \textbf n_{i}) \right\rbrace\) be a set of oriented point cloud
sampled from the surface of a solid \(M\), where \(\textbf c_{i}\in\mathbb{R}^{3}\)
denotes a spatial coordinate on the surface of the solid and \(\textbf
    n_{i}\in\mathbb{R}^{3}\) is its corresponding surface normal. Let \(\chi:\mathbb{R}^{3}\to\mathbb{R}\)
be the characteristic function. Then, the Divergence theorem states that
\[\iiint_{M} (\Delta\cdot\chi) dV = \iint_{\partial M}(\nabla\chi\cdot\textbf n) dS.\]
Approximating the right hand side of the above equation with the given
samples gives rise to the Poisson equation as follows:
\[\Delta \chi = \nabla \cdot \textbf v\]
where, \(\textbf v(\textbf x) = \sum_{(\textbf c_{i}, \textbf n_{i})\in\mathcal{P}}\delta(\textbf x - \textbf
    c_{i})\textbf n_{i}\) where \(\delta\) is the Kronecker delta. Solving this set of
linear Partial Differential Equations (PDEs) differentiably, involves
discretizing the point normal field \(\textbf v\) uniformly by rasterizing the
point normals onto a uniformly sampled voxel grid. The differentiability in
the point rasterization process comes via inverse trilinear
interpolation. With Sprectral methods, the original signal can be decomposed
into a linear sum of sine/cosine basis functions whose derivatives can be
computed analytically. Therefore, employing this method one can solve for
one can first solve for the unnormalized characteristic function \(\chi'\)
without the boundary conditions
\[\chi'=\operatorname{IFFT}(\tilde X),
    \quad
    \tilde X=\tilde g_{\sigma,r}(\textbf u)\odot \frac{i\textbf u\cdot\tilde{\textbf v}}{-2\pi\left\| u \right\|^{2}},
    \quad
    \tilde g_{\sigma,r}(\textbf u)=\exp\left( -\frac{2\sigma^{2}\left\| u \right\|^{2}}{r^{2}} \right)\]
where the Fast Fourier Transform of \(\textbf v\) is denoted as \(\tilde{\textbf
    v}=\operatorname{FFT}(\textbf v)\); \(\textbf u:=(u,v,w)\in\mathbb{R}^{n\times d}\) denotes the spectral
frequencies corresponding to the \((x,y,z)\) spatial dimensions and
\(\operatorname{IFFT}(\tilde \chi)\) represents the Inverse Fast Fourier Transform of
\(\tilde\chi\). \(\tilde g_{\sigma, r}(\textbf u)\) is a Gaussian smoothing kernel of
bandwidth \(\sigma\) at grid resolution \(r\) in the spectral domain to mitigate
ringing effects as a result of the Gibbs phenomenon from rasterizing the
point normals. Therefore the normalized differentiable characteristic
function is given by
\[\chi=\frac m{\operatorname{abs}(\chi'|_{\textbf x=0})}\left( \chi'-\frac1{\left| \left\lbrace \textbf
    c_{i} \right\rbrace \right|}\sum_{\textbf c\in\left\lbrace \textbf c_{i} \right\rbrace}\chi'|_{\textbf x=\textbf c} \right).\]
\subsubsection{Architechture}
\label{sec:orgc5979d4}

The very first component of the learning based pipeline is the convolutional
point encoder network as proposed in \cite{Peng2020}. This network encode the noisy
unoriented point cloud coordinates \(\left\lbrace \textbf c_{i} \right\rbrace\) into a feature
\(\phi_{\theta}\) encapsulating both local and global information about the input
point cloud. Here, \(\theta\) refers to network parameters.

Let \(\phi_{\theta}(\textbf c)\) denote the feature at any particular point \(\textbf c\)
obtained from feature volume \(\phi_{\theta}\) using trilinear
interpolation. Then given the feature \(\phi_{\theta}(\textbf c)\) a shallow
Multi-Layer Perceptron (MLP) \(\textbf f_{\theta}\) predict \(k\)-offsets for \(\textbf
    c\):
\[\Delta\textbf c=\textbf f_{\theta}(\textbf c,\phi_{\theta}(\textbf c))\]
Therefore we get the updated point positions \(\hat{\textbf c}\) by adding the
offsets \(\Delta\textbf c\) to the input point position \(\textbf c\). These additional
offsets densify the point cloud, leading to enhanced reconstruction
quality. Following the authors, we also conducted all our subsequent
experiments with \(k=7\).

Given the updated points \(\hat{\textbf c}\) a second MLP \(\textbf g_{\theta}\) is
trained to predict the corresponding normals: \[\hat{\textbf n}=\textbf
    g_{\theta}(\hat{\textbf c}, \phi_{\theta}(\hat{\textbf c})).\] The same decoder architecture
as in \cite{Peng2020} is used for both \(\textbf f_{\theta}\) and \(\textbf
    g_{\theta}\). The network comprises 5 layers of ResNet blocks with a hidden
dimension of 32.

\subsubsection{Training and Inference \label{org0b5cf8a}}
\label{sec:org45cebfb}
The authors used watertight and noise-free meshes for supervision and
acquire the ground truth indicator grid by running PSR algorithm
\cite{Kazhdan2006} on a densely sampled point clouds of the ground truth
meshes with the corresponding ground truth normals. Mean Square Error (MSE)
loss on the predicted \((\hat\chi)\) and ground truth \((\chi)\) indicator grid
\[\mathcal{L}_{\text{DPSR}}=\left\| \hat\chi-\chi \right\|^{2}\] is used for training the model with
a learning rate of 5e-4. During inference, the trained model predicts the
normals and offsets and then DPSR solves for the PSR indicator grid,
and after that Marching Cubes \cite{Lorensen1987} extract meshes from the
PSR indicator grid as demonstrated in Figure \ref{sap_pipeline}.

\subsection{Limitation of the Learning-based Pipeline in SAP}
\label{sec:org48f633f}
The limitation of the learning-based pipeline, for non-watertight mesh
reconstruction comes from the PSR algorithm itself.

By definition, the characteristic function relies on the assumption that a
solid object has an interior and an exterior part with an enclosed
boundary. In practice, deriving the functional form of the boundary of a
solid is very difficult and using it to solve for the characteristic function
is impractical. Thus, we resort to using the point cloud sampled from the
surface of the solid and solve the Poisson equation to get the PSR indicator
grid. But by using point cloud, we no longer have an enclosed boundary but
many tiny holes where no samples are present. In these regions where no
samples are present the indicator values diffuse from 1 to 0 gradually rather
than a sharp change. This problem can be partially mitigated in SAP by
densifying the point cloud by predicting offsets.

This problem is more pronounced in the case of non-watertight mesh
reconstruction since the regions where there are no samples are much
larger. Therefore, running the PSR algorithm on a point cloud sampled from a
non-watertight mesh results in a PSR indicator grid where a sharp transition
from 1 to 0 can be observed where there are samples (sharpness depends on the
point cloud density at that region) and regions where there are no samples,
indicator values gradually diffuse from 1 to 0. Therefore running Marching
Cube algorithm on this PSR grid results in a watertight mesh where
reconstructed surfaces are well constructed and match with the boundaries
defined by the samples but also produces excess surfaces of arbitrary
topology where there are no samples.

In this project, we tried to mitigate this problem by identifying the regions
with a sharp transition from 1 to 0 in the PSR grid.

\section{Our Approach \label{orga7e8a2d}}
\label{sec:org5c98b0c}
Given a noisy, unoriented point cloud in \(\mathbb{R}^{3}\) our goal is to reconstruct a
surface (watertight or non-watertight) that fits the point cloud. We do this by
extending the existing learning-based watertight surface reconstruction pipeline
in SAP \cite{peng2021shape} with a Surface Mask Prediction Network.

\subsection{Method}
\label{sec:orgbce455b}
To begin with, we define ``A Surface Mask" of an object surface to be a
volume in the 3D space that encapsulates the object surface. Given a PSR
indicator grid, our objective is to predict a surface mask such that it only
encapsulates the actual object surface and does not encapsulate other
non-surface region. Thus, our surface mask has to be thick enough such that
it captures the actual surface entirely and thin enough such that it does not
capture any non-surface region. The reason behind this approach is that once
we can generate the appropriate surface mask, we can run the marching cube
algorithm restricted to the masked region and therefore we can extract only
the relevant object surface.

To achieve this we observed that, the PSR indicator grid constructed from a
point cloud sampled from a non-watertight mesh shows a sharp change in the
gradient where there are point samples, whereas the gradient changes slowly
where there are no samples. But the sharpness of the gradient change depends
on the density of the point cloud in that region. Therefore, simple
hand-engineered gradient-based filtering is not sufficient to detect the
regions (See the comparisons below in Section \ref{orgd9ef9db}). Thus, we resort to
Machine Learning based approach.

\subsubsection*{Surface Mask Prediction Network}
\label{sec:org5cfc8ad}
We can regard the aforementioned problem as a semantic segmentation problem
in 3D as we want to label each voxel if it belongs in the surface mask
region and thus provides a dense 3D surface mask. Now the surface mask is a
local feature of the PSR indicator grid and requires a small receptive field
to label a voxel by identifying whether a sharp gradient change occurs in
the vicinity of the voxel. Therefore a standard 3D U-NET, with one input and
output channel, is sufficient to predict the surface mask.

\begin{figure}[h!]
\centering
\includegraphics[width=\textwidth]{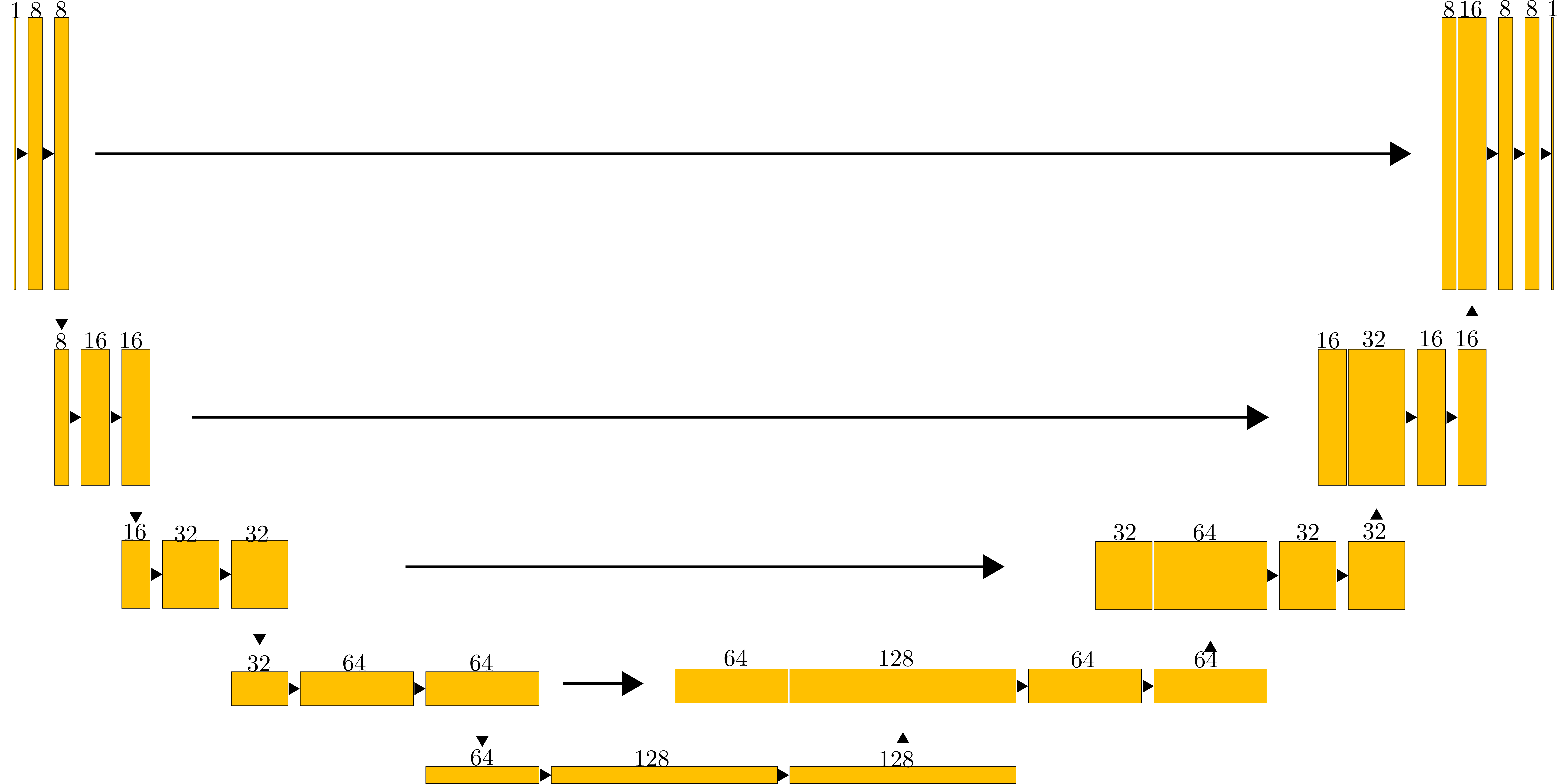}
\caption{Model architechture of the Surface Mask Prediction. A standard 3D U-NET architechture with feature channels 8, 16, 32, 64, 128.}
\label{architechture}
\end{figure}

Following the work on 3D U-NET \cite{Cicek2016}, in the analysis part, we
used at each stage a \texttt{DoubleConv} module consists of two \(3\times 3\times 3\)
convolutions each followed by batch normalization and a rectified linear
unit (ReLU) and then a \(2\times 2\times 2\) max-pooling with a stride of two in
each dimension. In the synthesis part, we used 3D transposed convolution
operator \cite{Zeiler2010} each followed by the \texttt{DoubleConv} module. We also
used skip-connections to fuse high-resolution features from the analysis
path into the synthesis path. We call this network ``Surface Mask Prediction
Network" (SMPN). The illustration of the network is given in Figure
\ref{architechture}.

\begin{figure}[h!]
\centering
\includegraphics[width=\textwidth]{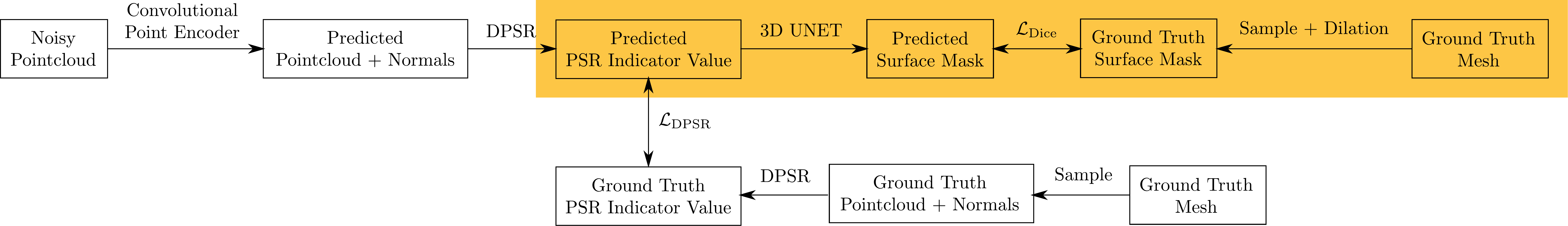}
\caption{Schematic view of the extended SAP pipeline. The highlighted portion indicates the extension.}
\label{pipeline}
\end{figure}

Now with the SMPN, we can extend the learning-based watertight surface
reconstruction pipeline as illustrated in Figure \ref{pipeline} to
facilitate non-watertight surface reconstruction.

\subsection{Implementation Details}
\label{sec:org59c633b}

\subsubsection{Dataset}
\label{sec:orgb5be1b9}
We generate our training data using the watertight meshes from ShapeNet
\cite{Chang2015a} dataset. For all our experiments we used the meshes of the
following classes --- Box, Car, Chair, Ship, and Sofa. 

\subsubsection{Data Preprocessing Pipeline \label{org82a1b3e}}
\label{sec:org47b3a92}
Given a watertight mesh from ShapeNet dataset \cite{Chang2015a}, We first
apply a random rotation to the mesh and then we translate and scale it to
fit into the \([0,1]^{3}\) unit cube. We then use this transformed mesh to
densely sample oriented points and feed them to the PSR algorithm
\cite{Kazhdan2006} to generate the ground truth PSR indicator grid. From
this point cloud, we then remove points that fall within some randomly
selected regions, which we use as the ground truth point cloud. We then
scale this ground truth point cloud and discretize it to create the voxel
grid and apply a \(7\times7\times7\) 3D dilation kernel on this voxel grid to
generate the ground truth surface mask.

\subsubsection{Training and Inference}
\label{sec:org0f7ac82}

During the training phase, we train both the learning-based watertight
surface reconstruction pipeline and the surface mask prediction network
jointly. We first obtain the predicted PSR indicator grid \(\hat\chi\) by
feeding the unoriented point cloud \(\hat{\textbf c}\) as input to the
learning-based pipeline in SAP and then use the predicted PSR indicator grid
\(\hat \chi\) as input to our Surface Mask Prediction Network to generate the
surface mask \(\hat M\). The corresponding ground truth mask \(M\) is
obtained as described in Section \ref{org82a1b3e}. To compute the loss between the
predicted and the ground truth mask, we used the dice loss
\cite{Milletari2016} \[\mathcal{L}_{\text{Dice}} = 1 - \frac{2\sum_{i,j,k}M_{ijk}\hat
    M_{ijk}+1}{\sum_{i,j,k}M_{ijk}^{2}+\sum_{i,j,k}\hat M_{ijk}^{2}+1}\] which
considers the loss information both locally and globally as it is critical
to predict a precise surface mask. Also to train the SAP pipeline we use the
same \(\mathcal{L}_{\text{DPSR}}\) loss as discussed in Section \ref{org0b5cf8a}. Thus to
train both the networks jointly, we used the following total loss
\[\mathcal{L}=\mathcal{L}_{\text{DPSR}}+\mathcal{L}_{\text{Dice}}.\] We implement all models in PyTorch
\cite{NEURIPS2019_9015} and for training, we use the Adam optimizer
\cite{Kingma2014} with a learning rate of 5e-4 and batch size of 16. We
trained the models using NVIDIA GeForce RTX 3090.

\begin{figure}[h!]
\centering
\includegraphics[width=\textwidth]{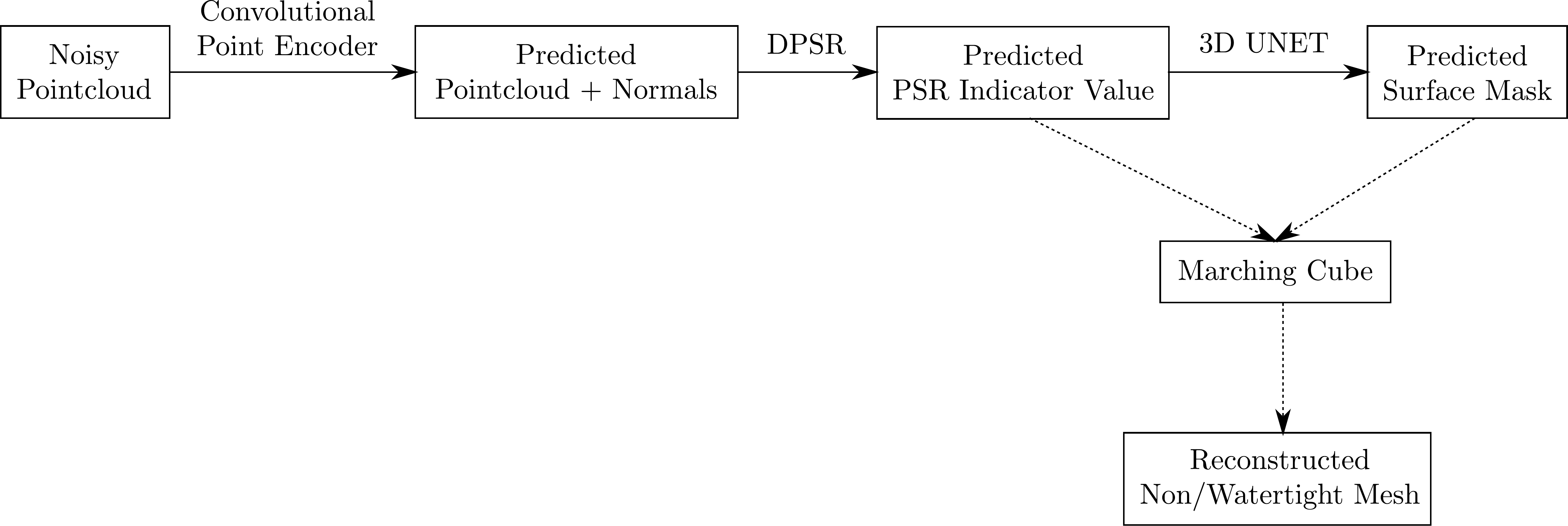}
\caption{Schematic view of the inference mechanism}
\label{sapeinference}
\end{figure}

During inference, given any input point cloud, we first use the
learning-based watertight surface reconstruction pipeline in SAP to predict
the PSR indicator grid and use the predicted indicator grid as input to our
surface mask prediction network to predict the surface mask and run Marching
Cubes \cite{Lorensen1987a} on the indicator grid restricted to the surface
mask region to extract the non-watertight mesh. Figure \ref{sapeinference}
illustrates the inference mechanism.

\section{Experimental Results \label{orge45565e}}
\label{sec:org293b2dc}

\subsection{Evaluation Metrics \label{orge630a53}}
\label{sec:org848116b}
We considered Chamfer Distance and Hausdorff Distance as our evaluation
metrics. Given two point clouds \(S_{1}\) and \(S_{2}\), the Chamfer
Distance (CD) and Hausdorff Distance (HD) is defined as follows
\[\operatorname{CD}(S_{1},S_{2})=\frac1{|S_{1}|}\sum_{x\in S_{1}}\min_{y\in
    S_{2}}\left\| x-y \right\|^{2}_{2} + \frac1{|S_{2}|}\sum_{x\in S_{2}}\min_{y\in
    S_{1}}\left\| x-y \right\|^{2}_{2},\] \[\operatorname{HD}(S_{1}, S_{2}) = \max\left\lbrace \sup_{x\in
    S_{1}}\inf_{y\in S_{2}}\left\| x-y \right\|, \sup_{y\in S_{2}}\inf_{x\in
    S_{1}}\left\| x-y \right\| \right\rbrace.\]

These metrics may not be the best indicator of similarity between two
non-watertight meshes since if one of the non-watertight meshes have some
tiny holes and the other does not, then this discrepancy may not get
reflected in the metric score as we are calculating the metric only with a
number of points sampled from the surface of the meshes. Therefore, we also
use visual aid for our assessment.

\subsection{Non-watertight Surface Reconstruction}
\label{sec:orgc450a96}
In this part, we investigate whether our extension to the SAP pipeline can
be used for non-watertight surface reconstruction from unoriented point
clouds. We evaluate our method on the single object reconstruction task
using noise and outlier-augmented point clouds from ShapeNet as input to our
method. We investigate the performance for three different noise levels: (a)
Gaussian noise with zero mean and standard deviation 0.005, (b) Gaussian
noise with zero mean and standard deviation 0.025, (c) 50\% points have the
same noise as in a) and the other 50\% points are outliers uniformly sampled
inside the unit cube. Table \ref{table1} shows that our method achieves
better results compared to the base SAP pipeline. Also, we can observe that
training the entire extended pipeline performs better than training only the
SMPN with the backbone frozen.  Table \ref{table2} shows qualitatively that
our method works well with various noise levels augmented in the input point
cloud.

\begin{center}
\begin{longtable}{|m{0.31\textwidth}| m{0.31\textwidth} | m{0.31\textwidth}|}
\hline
{\bfseries Pipeline} &
\raggedleft{\bfseries Chamfer Distance ($\downarrow$)} &
\raggedleft\arraybslash{\bfseries Hausdorff Distance ($\downarrow$)}\\\hline\hline
SAP &
\raggedleft 0.009798 &
\raggedleft\arraybslash 0.254690\\\hline
SAP(freezed) + SMPN &
\raggedleft 0.000615 &
\raggedleft\arraybslash 0.108691\\\hline
\cellcolor{SpringGreen!50}SAP + SMPN &
\cellcolor{SpringGreen!50} \raggedleft{\bfseries 0.000239} &
\cellcolor{SpringGreen!50} \raggedleft\arraybslash{\bfseries 0.071070}\\\hline
\caption{Quantitative comparison of the performances of the extended pipeline (2nd and 3rd row) with the base pipline (1st row) on the ShapeNet dataset (mean over 5 classes).}
\label{table1}
\end{longtable}
\end{center}
 \begin{center}
\begin{longtable}{m{0.05\textwidth}|m{0.2125\textwidth}|m{0.2125\textwidth}|m{0.2375\textwidth}|m{0.2125\textwidth}}
~
 &
\centering Input &
\centering SAP &
\centering SAP Extended &
\centering\arraybslash GT Mesh\\\hline &&&&\\
~
 &
\includegraphics[width=\linewidth]{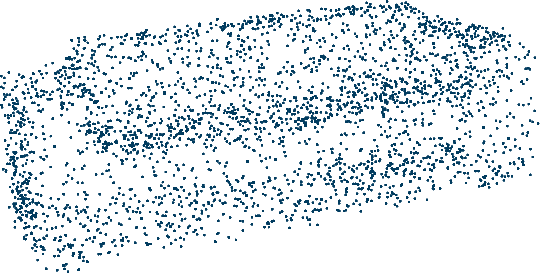} &
\includegraphics[width=\linewidth]{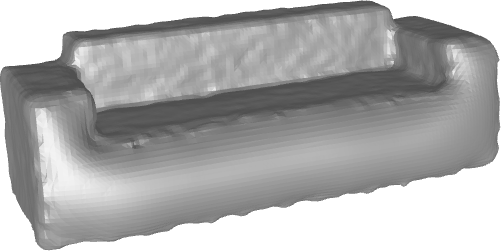} &
\includegraphics[width=\linewidth]{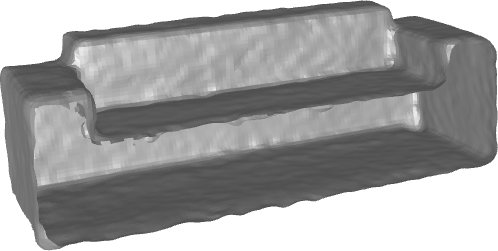} &
\includegraphics[width=\linewidth]{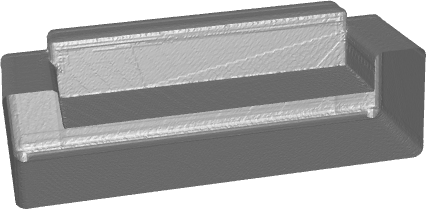}\\&&&&\\
\centering \rotatebox[origin=c]{90}{Low Noise} &
\includegraphics[width=\linewidth]{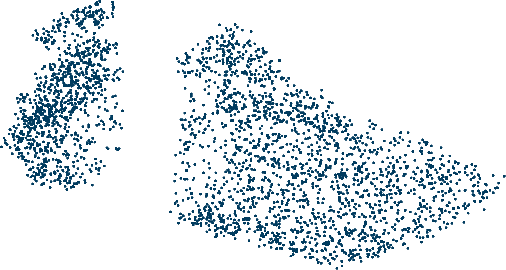} &
\includegraphics[width=\linewidth]{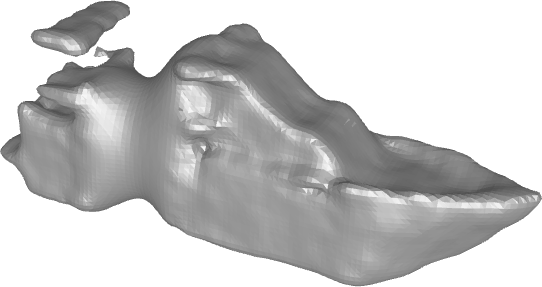} &
\includegraphics[width=\linewidth]{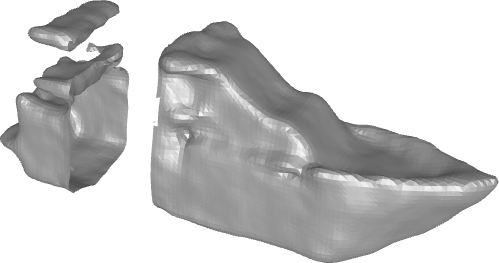} &
\includegraphics[width=\linewidth]{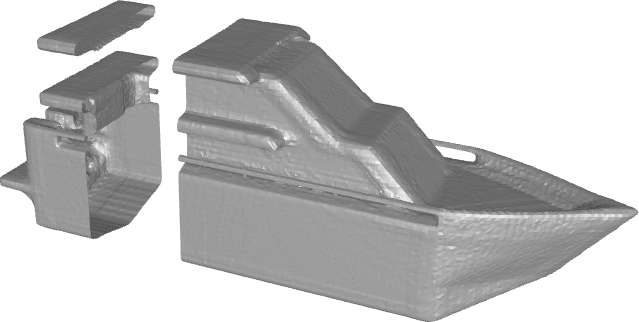}\\&&&&\\
~
 &
\includegraphics[width=\linewidth]{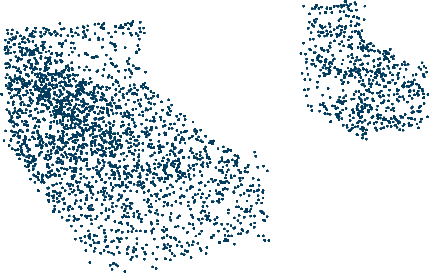} &
\includegraphics[width=\linewidth]{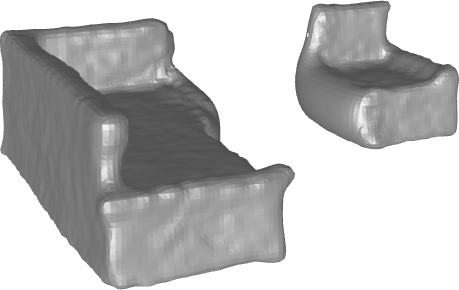} &
\includegraphics[width=\linewidth]{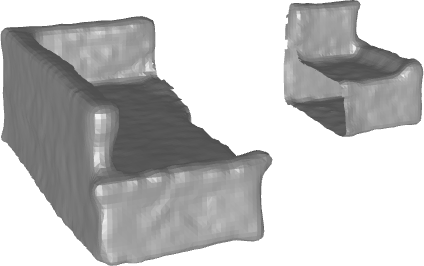} &
\includegraphics[width=\linewidth]{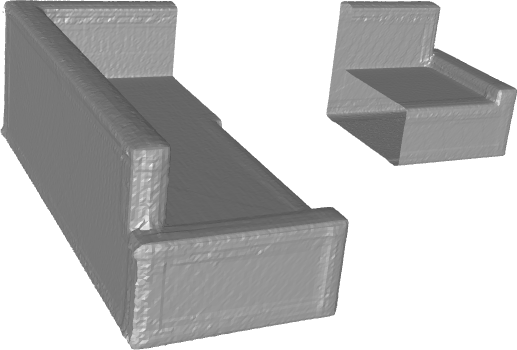}\\\hline &&&&\\
~
 &
\includegraphics[width=\linewidth]{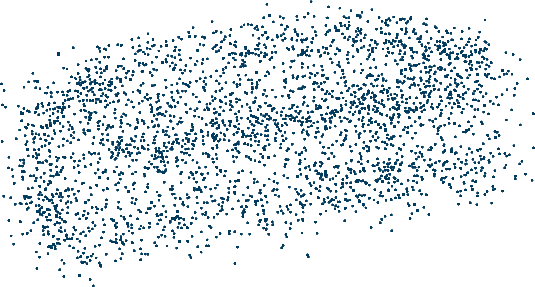} &
\includegraphics[width=\linewidth]{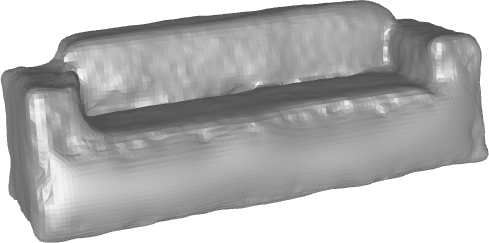} &
\includegraphics[width=\linewidth]{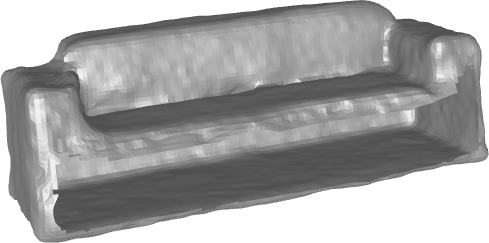} &
\includegraphics[width=\linewidth]{img/table1/1g.png}\\&&&&\\
\centering \rotatebox[origin=c]{90}{High Noise} &
\includegraphics[width=\linewidth]{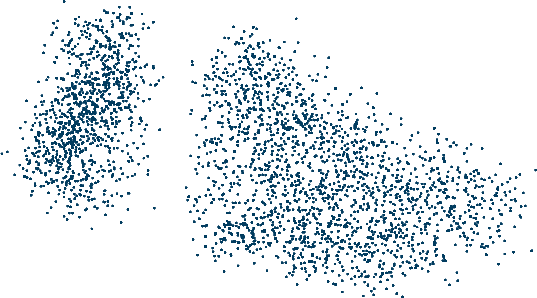} &
\includegraphics[width=\linewidth]{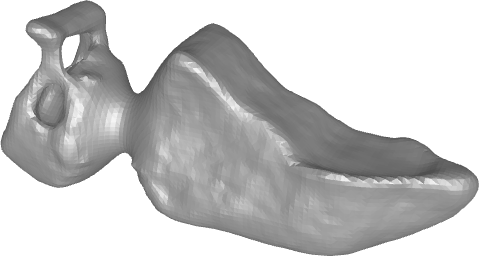} &
\includegraphics[width=\linewidth]{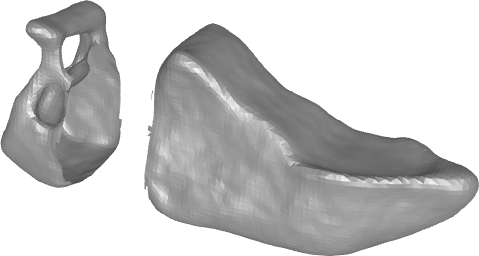} &
\includegraphics[width=\linewidth]{img/table1/2g.png}\\&&&&\\
~
 &
\includegraphics[width=\linewidth]{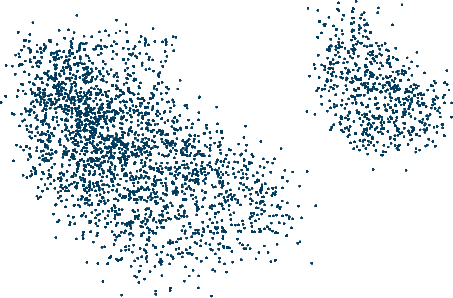} &
\includegraphics[width=\linewidth]{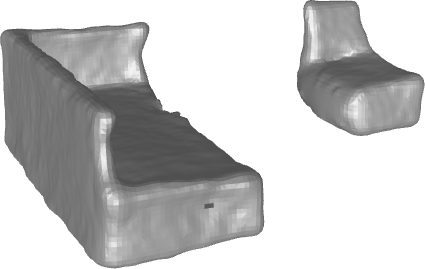} &
\includegraphics[width=\linewidth]{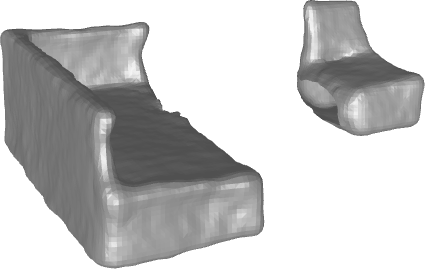} &
\includegraphics[width=\linewidth]{img/table1/3g.png}\\\hline&&&&\\
~
 &
\includegraphics[width=\linewidth]{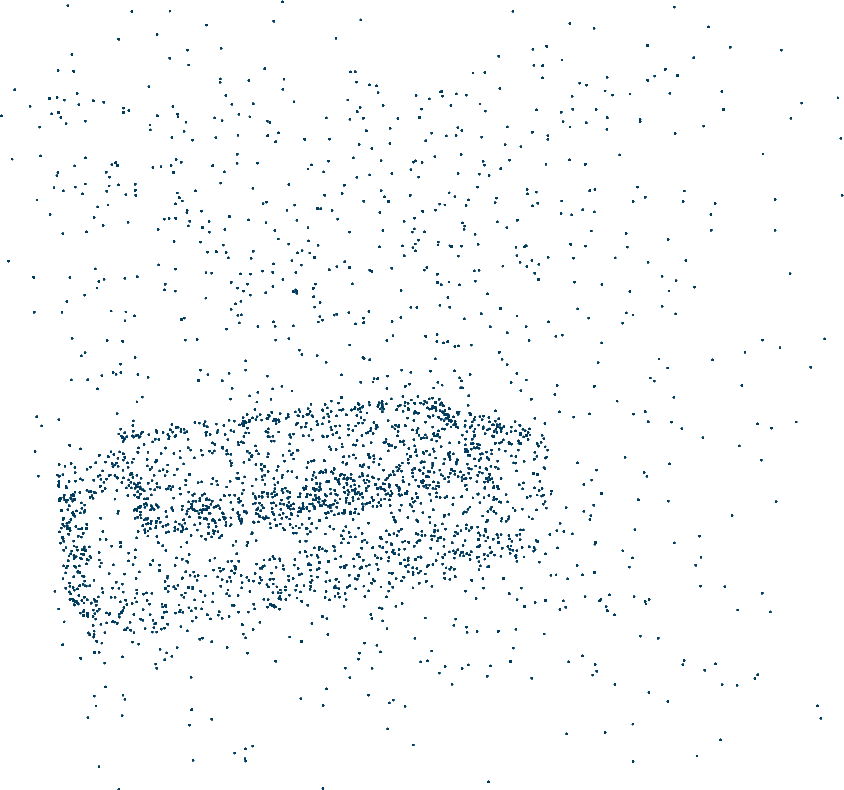} &
\includegraphics[width=\linewidth]{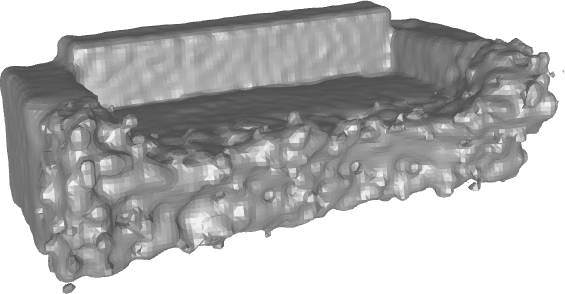} &
\includegraphics[width=\linewidth]{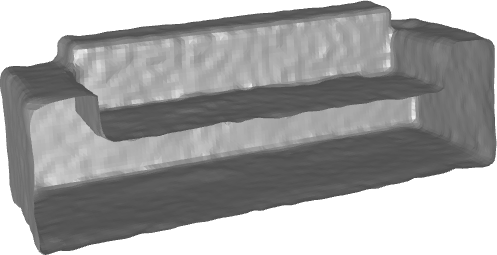} &
\includegraphics[width=\linewidth]{img/table1/1g.png}\\&&&&\\
\centering \rotatebox[origin=c]{90}{Outliers} &
\includegraphics[width=\linewidth]{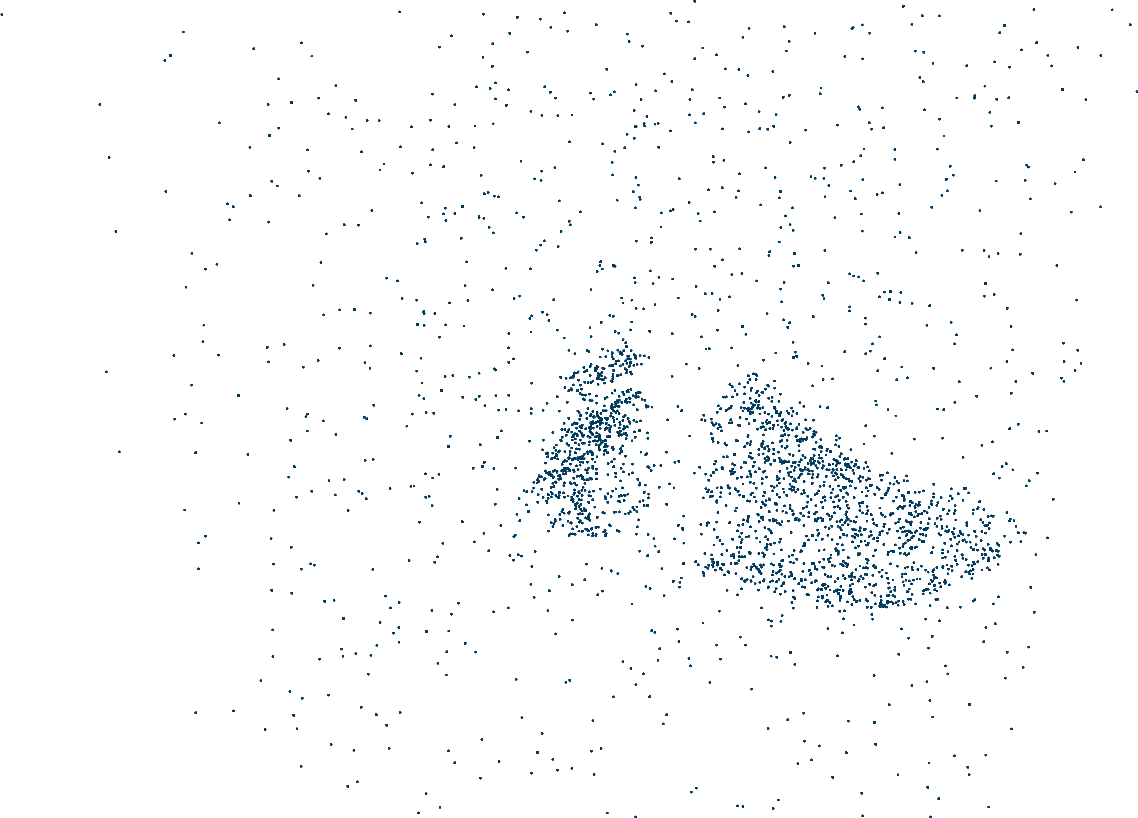} &
\includegraphics[width=\linewidth]{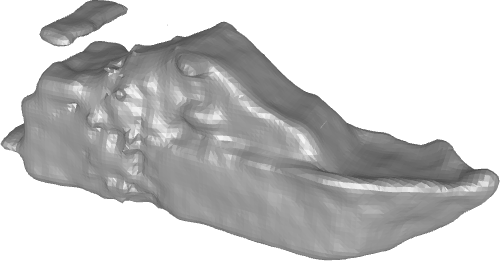} &
\includegraphics[width=\linewidth]{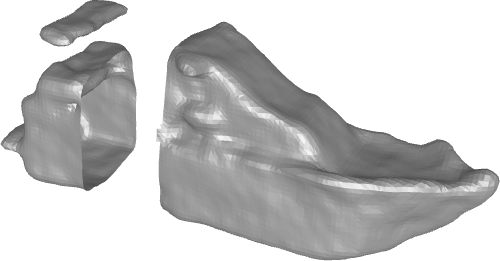} &
\includegraphics[width=\linewidth]{img/table1/2g.png}\\&&&&\\
~
 &
\includegraphics[width=\linewidth]{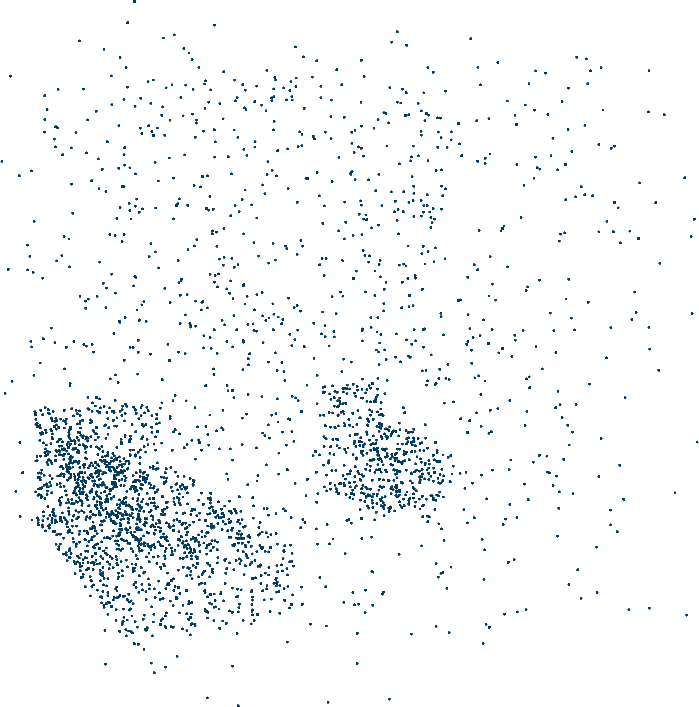} &
\includegraphics[width=\linewidth]{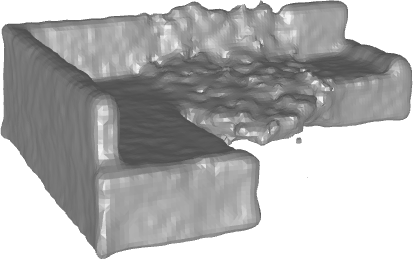} &
\includegraphics[width=\linewidth]{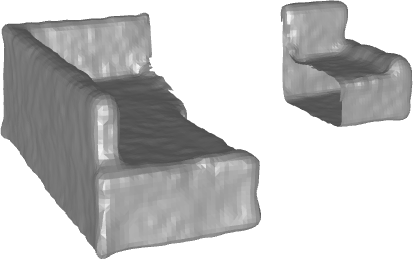} &
\includegraphics[width=\linewidth]{img/table1/3g.png}\\&&&&\\
\caption{Qualitative comparison of the performances of the extended pipeline with the base pipline in 3 different types of input.}
\label{table2}
\end{longtable}
\end{center}

\subsection{Comparisons to the Baselines Methods \label{orgd9ef9db}}
\label{sec:orgf667a1a}
For baselines, we used the classic Laplacian filter on the PSR indicator
grid with various thresholds for mask generation. We first discuss the
baseline method with 2D Laplacian filtering. Given a 3D PSR indicator grid,
we take each 2D slice and convolve it using a 2D \(3\times 3\) Laplace kernel:
\[\begin{bmatrix} 0&-1&0\\-1&4&-1\\0&-1&0 \end{bmatrix},\] to get a silhouette of the edges defined
by the indicator grid in that slice. Then we compute the absolute values of
the convolved grid and use different thresholds to classify whether a pixel
is an actual point on the surface of the mesh. Then, we apply a \(7\times 7\)
dilation kernel to generate the surface mask and use it together with the
PSR indicator grid to extract the non-watertight surface using Marching Cube
algorithm.

The baseline method for 3D Laplacian filter is similar as the 2D
baseline. In this case, we use a 3D \(3\times 3\times 3\) Laplace kernel \(K\) given
by \[K_1 = \begin{bmatrix} 0&0&0\\0&1&0\\0&0&0 \end{bmatrix},\quad K_2 =
    \begin{bmatrix} 0&1&0\\1&-6&1\\0&1&0 \end{bmatrix},\quad K_3 = \begin{bmatrix} 0&0&0\\0&1&0\\0&0&0 \end{bmatrix},\] where
\(K_{i}\) denotes the \(i\)-th plane and we use this kernel to convolve the
entire grid. Then we compute the absolute values of convolved grid and use
thresholding as above and convolve it using a \(7\times 7\times 7\) dilation kernel
to generate the surface mask.

Table \ref{table3} and Table \ref{table4} show the quantitative and
qualitative comparisons respectively. We can observe that the baseline
methods do not perform well. Also, a considerable output depends on the
threshold and varies from input to input. Compared to the baseline methods,
our method achieves superior performance. \begin{table}[H]
\begin{longtable}{|m{0.20\textwidth}|m{0.14\textwidth}|m{0.29\textwidth}|m{0.29\textwidth}|}
\hline
{\bfseries Method} &
\raggedleft{\bfseries Threshold} &
\raggedleft{\bfseries Chamfer Distance ($\downarrow$)}&
\raggedleft\arraybslash{\bfseries Hausdorff Distance ($\downarrow$)}\\\hline\hline
2D Laplacian &
\raggedleft 0.00 &
\raggedleft 0.009798 &
\raggedleft\arraybslash 0.254690\\\hhline{~---}
~
 &
\raggedleft {\bfseries 0.05} &
\raggedleft {\bfseries 0.008534} &
\raggedleft\arraybslash {\bfseries 0.254473}\\\hhline{~---}
~
 &
\raggedleft 0.10 &
\raggedleft 0.009470 &
\raggedleft\arraybslash 0.254429\\\hhline{~---}
~
 &
\raggedleft 0.20 &
\raggedleft 0.016016 &
\raggedleft\arraybslash 0.260365\\\hhline{~---}
~
 &
\raggedleft 0.40 &
\raggedleft 0.059645 &
\raggedleft\arraybslash 0.340960\\\hhline{----}
3D Laplacian &
\raggedleft 0.00 &
\raggedleft 0.009799 &
\raggedleft\arraybslash 0.254690\\\hhline{~---}
~
 &
\raggedleft {\bfseries 0.05} &
\raggedleft {\bfseries 0.008783} &
\raggedleft\arraybslash {\bfseries 0.254579}\\\hhline{~---}
~
 &
\raggedleft 0.10 &
\raggedleft 0.010009 &
\raggedleft\arraybslash 0.254576\\\hhline{~---}
~
 &
\raggedleft 0.20 &
\raggedleft 0.023703 &
\raggedleft\arraybslash 0.255712\\\hhline{~---}
~
 &
\raggedleft 0.40 &
\raggedleft 0.088625 &
\raggedleft\arraybslash 0.400718\\\hhline{----}\hline
\cellcolor{SpringGreen!50}Ours &
\cellcolor{SpringGreen!50} \raggedleft {}- &
\cellcolor{SpringGreen!50} \raggedleft{\bfseries 0.000239} &
\cellcolor{SpringGreen!50} \raggedleft\arraybslash{\bfseries 0.071070}\\\hline
\caption{Quantitative comparison of the performances of the extended pipeline with the baseline methods on the ShapeNet dataset (mean over 5 classes). The table shows performances of the kernel methods (2D/3D Laplacian) for mask generation at various thresholds.}
\label{table3}
\end{longtable}
\end{table}
\begin{table}[p]
\begin{longtable}{m{0.18\linewidth}|m{0.18\linewidth}|m{0.18\linewidth}|m{0.18\linewidth}|m{0.18\linewidth}}
\centering Input &
{\centering 2D Laplacian\par}

\centering (with best threshold) &
{\centering 3D Laplacian\par}

\centering (with best threshold) &
\centering SAP Extended &
\centering\arraybslash Ground Truth\\\hline\hline &&&&\\
\centering \includegraphics[width=\linewidth]{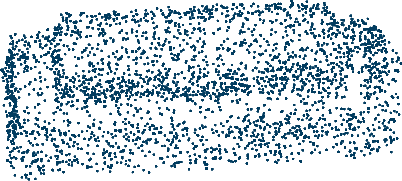} &
\centering \includegraphics[width=\linewidth]{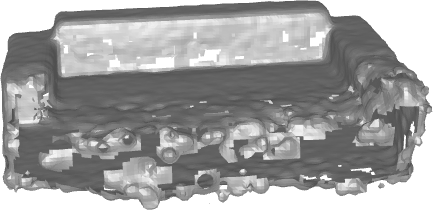} &
\centering \includegraphics[width=\linewidth]{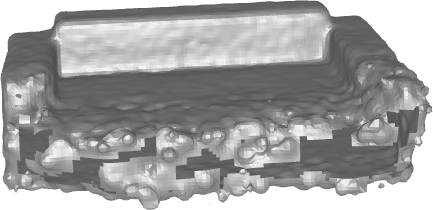} &
\centering \includegraphics[width=\linewidth]{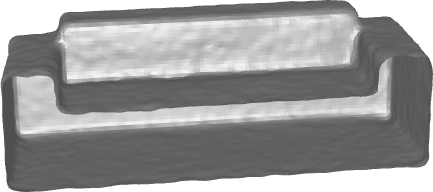} &
\centering\arraybslash \includegraphics[width=\linewidth]{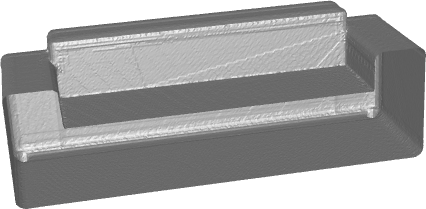}\\&&&&\\
\centering \includegraphics[width=\linewidth]{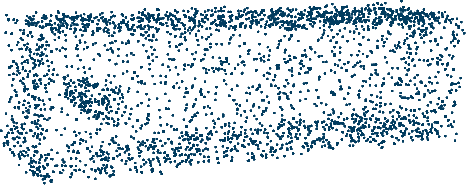} &
\centering \includegraphics[width=\linewidth]{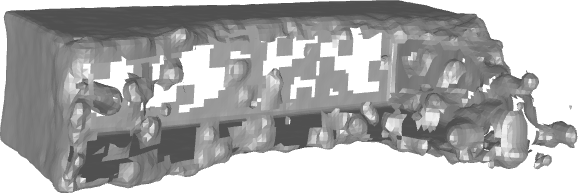} &
\centering \includegraphics[width=\linewidth]{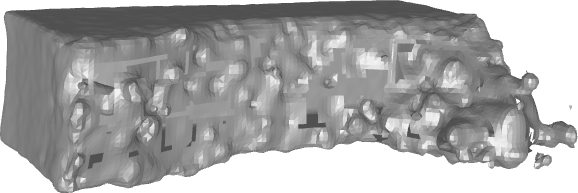} &
\centering \includegraphics[width=\linewidth]{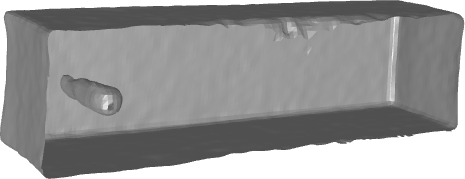} &
\centering\arraybslash \includegraphics[width=\linewidth]{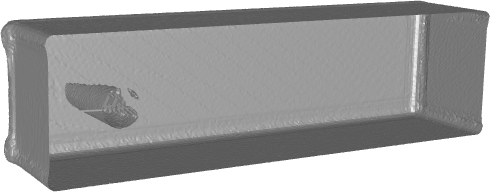}\\&&&&\\
\centering \includegraphics[width=\linewidth]{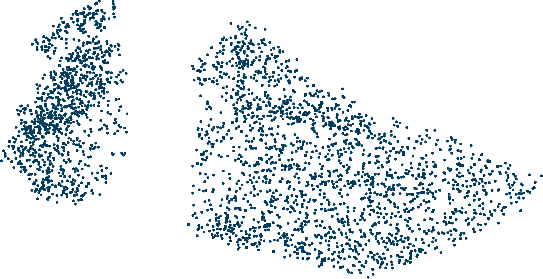} &
\centering \includegraphics[width=\linewidth]{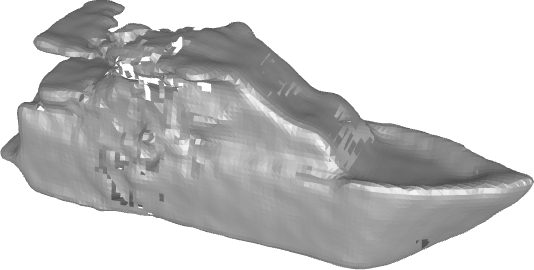} &
\centering \includegraphics[width=\linewidth]{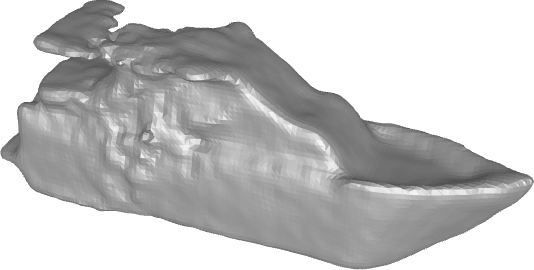} &
\centering \includegraphics[width=\linewidth]{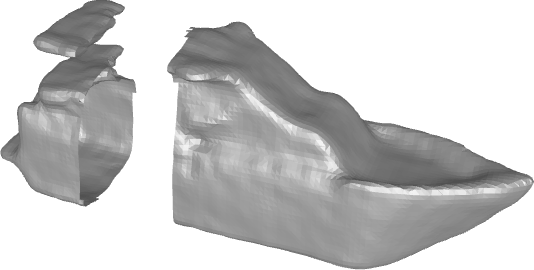} &
\centering\arraybslash \includegraphics[width=\linewidth]{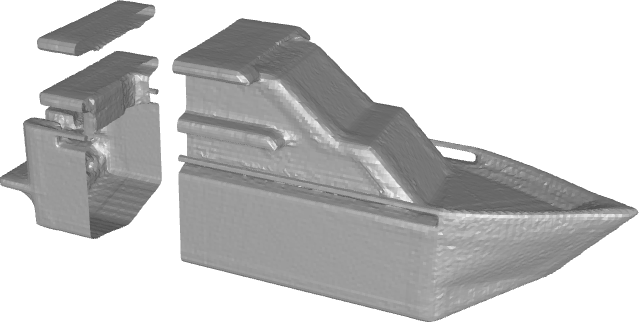}\\&&&&\\
\centering \includegraphics[width=\linewidth]{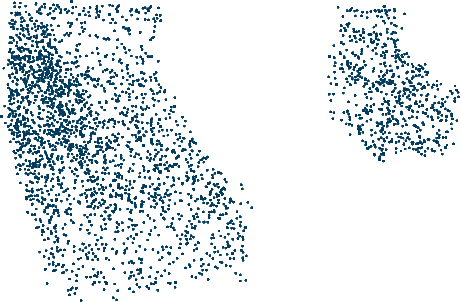} &
\centering \includegraphics[width=\linewidth]{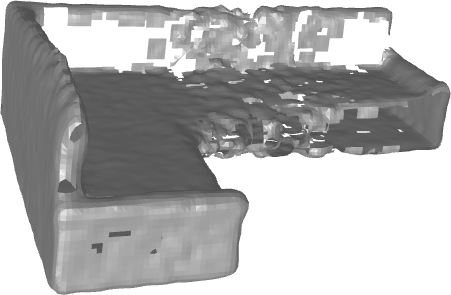} &
\centering \includegraphics[width=\linewidth]{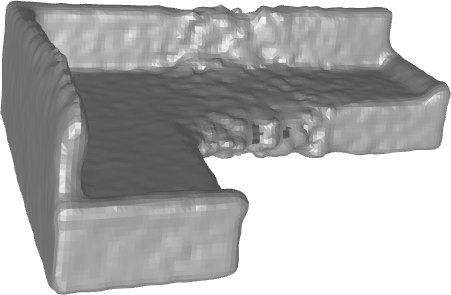} &
\centering \includegraphics[width=\linewidth]{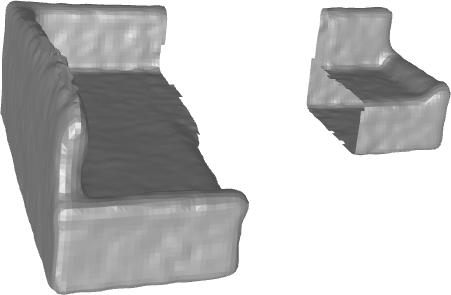} &
\centering\arraybslash \includegraphics[width=\linewidth]{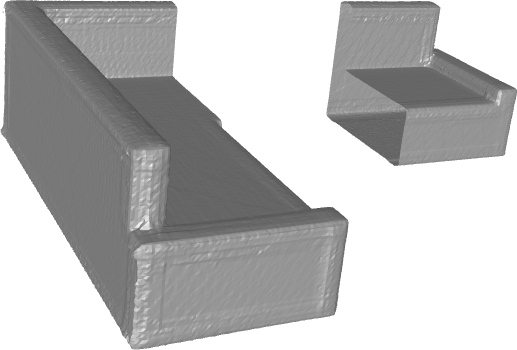}\\&&&&\\
\centering \includegraphics[width=\linewidth]{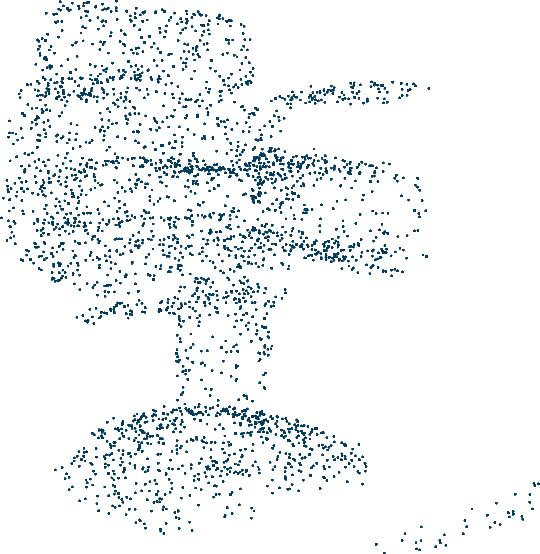} &
\centering \includegraphics[width=\linewidth]{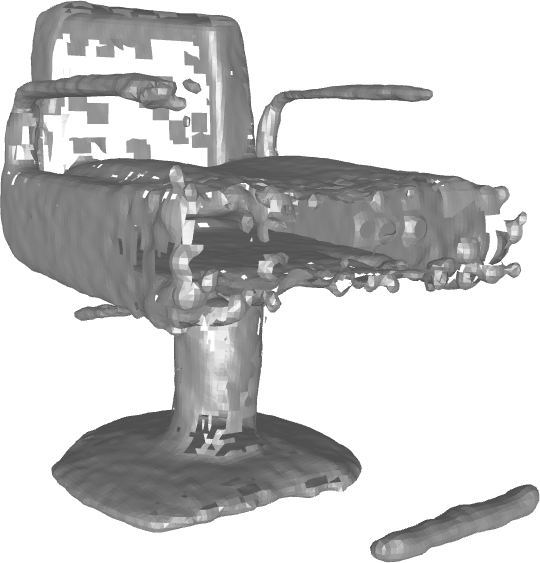} &
\centering \includegraphics[width=\linewidth]{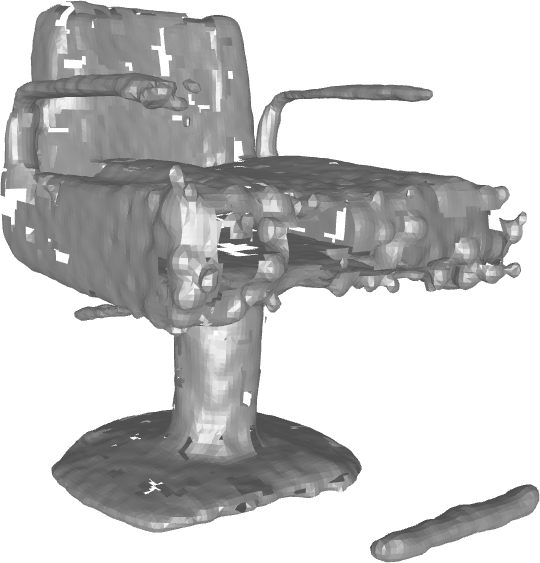} &
\centering \includegraphics[width=\linewidth]{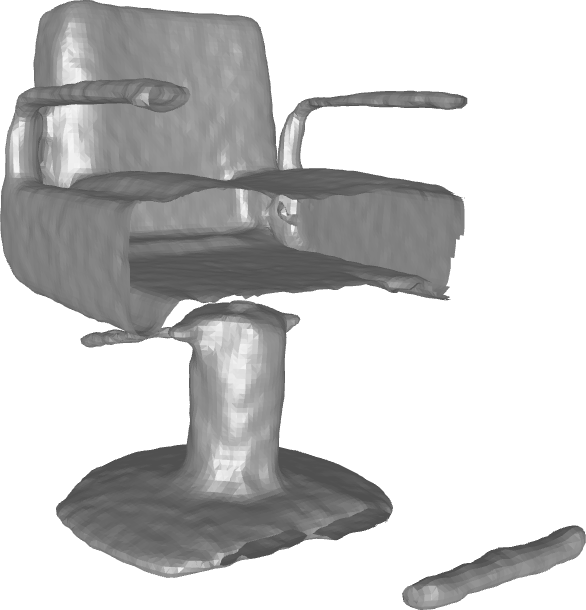} &
\centering\arraybslash \includegraphics[width=\linewidth]{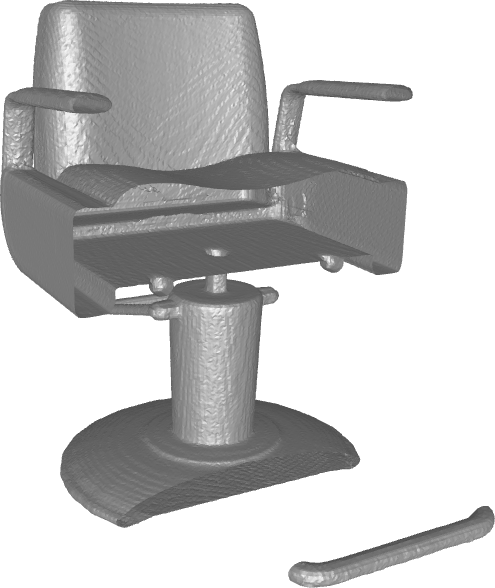}\\
\caption{Qualitative comparison of the performances of the extended pipeline with the baseline methods (with the best threshold for mask generation) on the ShapeNet dataset.}
\label{table4}
\end{longtable}
\end{table}

\subsection{Ablation Studies}
\label{sec:org694248a}
In this section, we investigate the influence of different dilation kernels
on the surface mask. We conduct our ablation experiments on ShapeNet for the
first setup i.e. with small induced noise (otherwise metric calculation
would be inconclusive). We generate our training data with dilation kernels
of various widths. We keep the Surface Mask Prediction Network and the SAP
pipeline fixed with their optimal configurations. We then train the model
with training data generated with a particular dilation kernel and then
measured its performance. Table \ref{table5} reports the results.
\begin{table}[H]
\begin{longtable}{|l|r|r|r|r|}
\hline
\diagbox{Metric}{Kernel Width} &
\raggedleft 3 &
\raggedleft 5 &
\raggedleft 7 &
\raggedleft\arraybslash 9\\\hline
Chamfer Distance &
0.000172
 &
\textbf{0.000174}
 &
0.000221
 &
0.000263
\\\hline
Hausdorff Distance &
0.047733
 &
\textbf{0.042084}
 &
0.060259
 &
0.058211
\\\hline
\caption{Ablation study over kernel width}
\label{table5}
\end{longtable}
\end{table} As we have pointed out before (in Section \ref{orge630a53}),
even though the metric shows better performance with kernel width 3, it is
not an accurate assessment. We found that the predicted mesh contains very
tiny holes at undesired places as being very thin, and the predicted surface
mask could not capture the regions where the surface lies very
accurately. On the other hand, we notice that training data generated with
kernel width 5 gives the best prediction performance both qualitatively and
quantitatively.

\section{Conclusion}
\label{sec:org93ea696}

In this project, we have presented a novel method for non-watertight mesh
reconstruction performing semantic segmentation on the PSR indicator grid. We
demonstrate its effectiveness in reconstructing the non-watertight mesh,
compared to the filtering-based baseline methods, both quantitatively and
qualitatively.

\textbf{Limitations.} The main limitation of our approach is the usage of dilation
kernels to generate the ground truth surface mask. Therefore, it will be
interesting to study if there is any way to predict the surface mask without the
supervised training. Secondly, our experiments were restricted to reconstructing
a single object surface. But we believe that our method can be extended to
reconstruct large scenes by combining small non-watertight surface patches,
reconstructed in a sliding-window manner. Finally, our initial approach is not
end-to-end. Therefore, taking supervision directly from the non-watertight mesh
via chamfer loss can be an interesting end-to-end approach to consider for
future studies.

\section{Acknowledgements}
\label{sec:orge30ce62}

I want to thank Songyou Peng and Chiyu `Max' Jiang for supervising the project
and providing me with the necessary guidance and valuable support throughout
this research project. I want to thank Prof. Andreas Geiger for the helpful
discussions. Also, I want to thank Songyou Peng and Madhav Iyengar for
proofreading the report.

\bibliography{sap}
\end{document}